\pdfoutput=1

\documentclass[11pt]{article}

\usepackage{mathtools}    

\DeclareMathOperator*{\argmax}{argmax}

\usepackage{acl}

\usepackage{times}
\usepackage{latexsym}

\usepackage[T1]{fontenc}

\usepackage[utf8]{inputenc}

\usepackage{microtype}

\title{Derivative Free Weight-space Ensembling}

\author{Dean Ninalga \\
  \texttt{justin.ninalga@mail.utoronto.ca} }

\begin{document}
\maketitle
\begin{abstract}
Recent work suggests that interpolating between the weights of two specialized language models can transfer knowledge between tasks in a way that multi-task learning cannot.
However, very few have explored interpolation between more than two models, where each has a distinct knowledge base.
In this paper, we introduce Derivative Free Weight-space Ensembling (DFWE), a new few-sample task transfer approach for open-domain dialogue. 
Our framework creates a set of diverse expert language models trained using a predefined set of source tasks. Next, we finetune each of the expert models on the target task, approaching the target task from several distinct knowledge bases. Finally, we linearly interpolate between the model weights using a gradient-free-optimization algorithm, to efficiently find a good interpolation weighting.
We demonstrate the effectiveness of the method on 
FETA-Friends \cite{albalak-etal-2022-feta} outperforming the standard pretrain-finetune approach.
\end{abstract}


\begin{figure*}[ht]
\centering
\includegraphics[width=\textwidth]{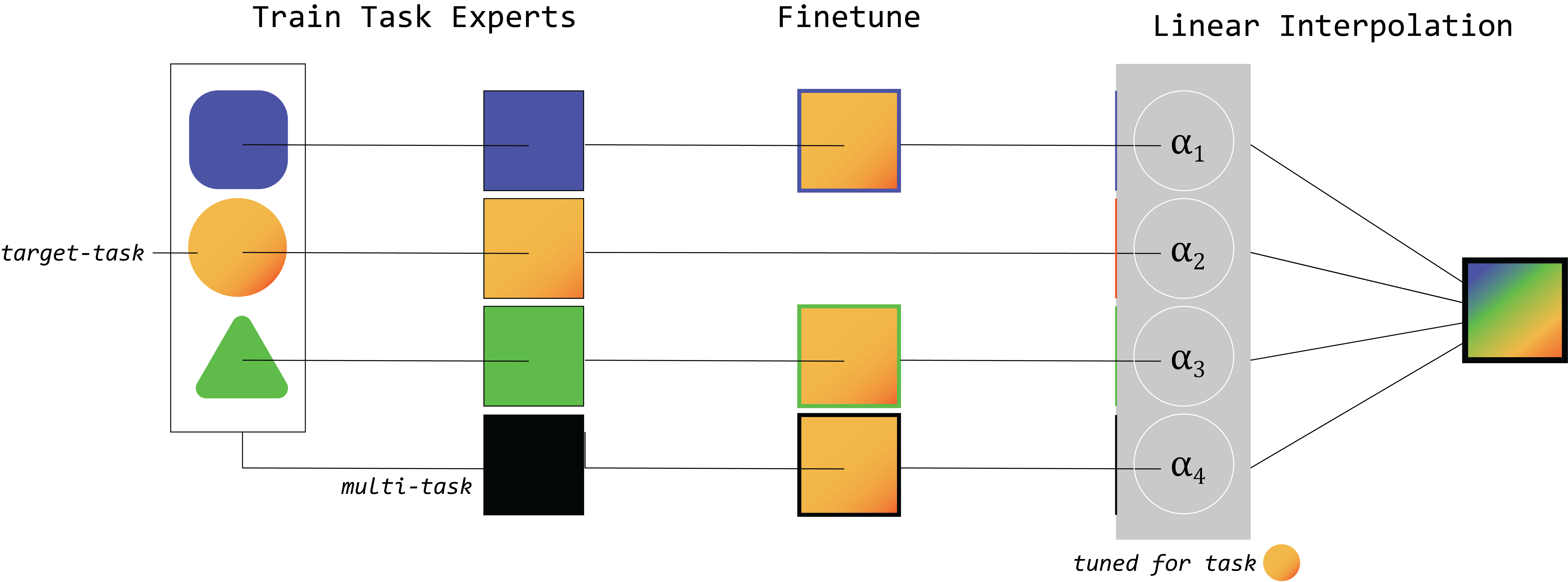}
\caption{\label{fig:ill}
An overview of our framework. First, we train task experts on a set of tasks (squares represent models). Next, we finetune each of the expert models on the target task, to diversify the fine-tunings. Finally, we linearly aggregate the resulting set of model weights into the final model. See Section \ref{sec:meth} for full details.}
\end{figure*}

\section{Introduction}
The growth in the number of open-vocabulary language models trained with multi-task learning has skyrocketed (as surveyed by \citet{Zhao2023ASO}).
They have had an integral part in the recent global success of large-scale conversational AI in recent years.
Open-vocabulary models typically demonstrate strong zero-shot performance and are trained on a never-ending list of source tasks.
For example, Flan-T5 \cite{https://doi.org/10.48550/arxiv.2210.11416}
is trained with 1800+ tasks and can be shown to have decent performance on unseen tasks.
However, recent work finds finetuning pre-trained models for specialized knowledge of a single informative task can yield better zero-shot performance than some of the latest zero-shot networks trained on hundreds or thousands of tasks \cite{Jang2023ExploringTB}.
Hence, avoiding non-generalizable knowledge contained in performing a set of source tasks, or the avoidance \emph{negative transfer}, remains an area of concern in transfer learning research \cite{Zhang2020ASO, Yu2022ASO}.

In the few-shot learning setting, recent work has shown that interpolating between the weights of a zero-shot and finetuned models can give superior performance to finetuning \cite{gagnon-audet2023awe}.
Indeed, recent studies suggest that even interpolating between two identical open-vocabulary language models that are trained on the same task can achieve even better performance \cite{Gueta2023KnowledgeIA}.
Here, we attempt to expand on these recent findings by using interpolation to distill knowledge from several models finetuned on a single task.
However, to the best of our knowledge, there is no known application of weight interpolation between more than \emph{two} task-expert models.
Hence, in order to optimize the interpolation of multiple model parameters - while maintaining a manageable computational cost - we propose derivative-free optimization to directly model the target task metric.

To diversify the weights from several models finetuned on a single task we adopt the tuning strategy of \emph{Model Ratatouille} \cite{Rame2022RecyclingDM}. Namely, given a set of $n$ auxiliary source tasks and a single target task, we first train $n$ new models for each of the Subsequently, we finetune each of the $s$ models on the target task.
However, unlike \emph{Model Ratatouille} \cite{Rame2022RecyclingDM} we do not perform linear probing in this work to reduce implementation complexity.

In this paper, our contributions are the following:
\begin{itemize}
    \item We describe Derivative Free Weight-space Ensembling (DFWE) our framework towards few-shot transfer for language understanding in open-domain dialogue. 
    \item We provide results on FETA-Friends 
\cite{albalak-etal-2022-feta} showing the effectiveness of our framework compared to standard approaches. 
\item We highlight limitations and directions for future improvements. 
\end{itemize}

\section{Related Work}
\subsection{Gradient-Free Optimization of Language Models}
Recently, several gradient-free optimization approaches have been proposed to tune extremely large language models that are prohibitively expensive to finetune with traditional differentiable network optimization \cite{Sun2022BlackBoxTF, Sun2022BBTv2TA, Han2023WhenGD}.
In broad terms, gradient-free approaches only use the forward pass of a network, making these approaches much more memory efficient and very attractive for tuning larger models.
In particular, \citet{Shen2023ReliableGA} demonstrates a gradient-free optimization approach capable of tuning a large language model only using the classification metric as feedback.

\subsection{Weight Interpolation}
ColD Fusion \cite{DonYehiya2022ColDFC} shows that a cycle of finetuning and weight averaging on a set of tasks can gradually outperform the multi-task learning approach on the same tasks.
However, ColD Fusion requires constant retraining on all the available tasks which can be very inefficient.
WiSE-FT \cite{Wortsman2021RobustFO} is a much more efficient framework that first proposed interpolating between the weights of a strong zero-shot model and the weights of fine-tuned models, leveraging the benefits of each approach.
Where \citet{Ilharco2022PatchingOM} recently verify the effectiveness of WiSE-FT with CLIP \cite{Radford2021LearningTV}.
Recently, AWE \cite{gagnon-audet2023awe} demonstrate WiSE-FT interpolation in the k-shot learning setting, where the interpolation parameter is determined using the data seen in the k-shots. Here, we will attempt to generalize WiSE-FT interpolation to more than two models by using weights from several fine-tuned models tuned on a single task.

\section{Preliminaries}
In this setting we are given a set of source tasks $S=\{s_1, s_2,...,s_n\}$ and a target task $t$. 
Each task $\tau\in\{t\}\cup S$ has a task specific metric $m_\tau(f^{(k)}(x_{\tau}),y_\tau)$, for network predictions $f^{(k)}(x_{\tau})$ after k-shots, input data $x_{\tau}$ and task labels $y_\tau$. Where the metric is not necessarily differentiable (e.g. F1, AUC, and, Acc).
Here our goal is to increase performance on a target task $t$ by using the knowledge obtained by performing the tasks in $S$. 
Namely, when measured on a test set the value of the metric $m_t$ using a model trained using $S$ must be greater than the value of $m_t$ using the outputs of a model not trained with $S$.

\section{Methodology}\label{sec:meth}

Here, we outline our method in greater detail.
We provide an illustration overview of the framework in Figure \ref{fig:ill}.
Our framework uses the T5-Flan-base\footnote{https://huggingface.co/google/flan-t5-base} model as it had better baseline performance over the T5 \cite{2020t5} model in our preliminary experiments.
Much of our framework builds upon the \emph{TLiDB
}\footnote{https://github.com/alon-albalak/TLiDB/tree/master} package, which is used to produce the baseline frameworks provided in \citet{albalak-etal-2022-feta}.
Here, we train using the same preprocessing, prompts, and instructions used for tuning the T5  model in \citet{albalak-etal-2022-feta}, since here, we train our T5-Flan-base models using the T5 training pipeline in the \emph{TLiDB} package.

\subsection{Baseline Model}
We will compare our transfer strategy to the standard transfer learning approach. That is, we finetune the pre-trained T5-Flan-base model on the target task.

\subsection{Our Approach}
Our training framework consists of three training stages. The first training stage gathers source-task-specific knowledge not present in the target task.
Where the second and third stages are designed to distill the knowledge gained from the first step for transfer to the target task.  

\subsubsection{Training Stage}
Instead of training a new model on each task in $S$, we instead construct $$S^{*}:=\{t,\{t\}\cup S,s_1, s_2,..,s_n\}.$$ Where we train a single model for each of the $n+2$ training sets in $S^*$. To be consistent with WiSE-FT \cite{Wortsman2021RobustFO}, we include training on the set $\{t\}\cup S$, which trains on all source tasks and the target task simultaneously. All models are trained until convergence on the metric for task $t$.

\subsection{Finetuning Stage}
Next, we take the resulting $|S^*|=n+2$ models and finetune them on the target task $t$.

\subsection{Interpolation Stage}
Subsequently, we use the resulting parameters from each model $\Theta:=\{\theta_{1}, ...,\theta_{|S^*|}\}$  to construct the final model: 
\begin{equation}\label{eq:param}
\theta(\Theta, A):=\sum_{i=1}^{| S^*|}\alpha_{i}\theta_{i}
\end{equation}
where for $A:=\{\alpha_i|\alpha_i \geq 0, \sum_{i}\alpha_{i}=1\}$. Here we use the Nelder-Mead algorithm \cite{Nelder1965ASM, Gao2012ImplementingTN} a gradient-free optimizer, for optimizing $A$ on the target task metric of the resulting model on the development data.

Formally, we optimize:
\begin{equation}\label{eq:astar}
    A^*:=\argmax_{A}m_{t}(f^{(k)}(x_{t}^{dev}|\theta(\Theta, A)),y_t)
\end{equation}

In practice, we run the optimizer for 40 iterations to find an estimate for $A^*$ which we then use to parameterize our final model (see Equation \ref{eq:param}) 
 used for inference.

\begin{table}
\centering
\begin{tabular}{l|l}
\hline
Parameter & Value \\
\hline
Number of Source-Tasks & 3\\ 
Optimizer & Adam \\
Learning Rate & 1e-4 \\
Float16 & False \\
Max Input Length & 512 \\
GPU Batch Size & 8 \\
Effective Batch Size & 24 \\
Epochs & 5\\
\hline
\end{tabular}
\caption{\label{tab:hyp}
Training Hyper-parameters
}
\end{table}

\begin{table*}
\centering
\begin{tabular}{l|lll}
\hline
\textbf{Task} & \textbf{Baseline Finetuning} & \textbf{DFWE (ours)} & \textbf{Score Delta} \\
\hline
\emph{emory emotion recognition} & 31.167 & 34.099 & 2.932\\
\emph{reading comprehension} & 57.093 & 58.888 & 1.795\\
\emph{character identification} & 62.366 & 64.469 & 2.103\\
\emph{question answering} & 39.758 & 41.027 & 1.27\\
\emph{personality detection} & 53.818 & 57.636 & 3.818\\
\emph{relation extraction} & 37.699 & 41.733 & 4.034\\
\emph{MELD emotion recognition} & 48.037 & 52.63 & 4.593\\
\hline
\end{tabular}
\caption{\label{tab:res}
Test Set Results on the FETA-Friends \cite{albalak-etal-2022-feta} task.
Scores are computed from task-specific evaluation metrics used for the target task (e.g. Accuracy for \emph{character identification}) where a score can be the average of several evaluation metrics.
}
\end{table*}

\section{Experimental Setup}
We study the transfer using our approach on the FETA-Friends \cite{albalak-etal-2022-feta} dataset, where there are several annotated tasks.
For each task $t$ in the dataset, we use the remaining tasks to construct the set of source tasks $S$.
We compare our results to the standard pretrain-finetune approach, where we finetune a pre-trained model on the target task.

We train all models using the hyper-parameters in Table \ref{tab:hyp}. 
Training using our framework on all tasks in the FETA-Friends dataset, including baseline fine-tunings,  takes about 3-5 days using a single T4 GPU.

\section{Results and Discussion}
Table \ref{tab:res} reports the test set results for each of the tasks in the FETA-Friends dataset.
Our run achieves a uniform improvement on the test set. Where the score delta, the improvement over the baseline standard finetuning method, is 2.935 on average.

Here, we do not perform any automated task selection and relied on hand-crafted source-task combinations.
Moreover, simply including all tasks in the set of source tasks was not as performant as restricting the number of source tasks to three.
Hence, future work may seek to incorporate an automated task-selection process that can regularize the number of included source tasks. 
A possible solution may include a source task weighting strategy, of which, there are several approaches designed for transfer learning setting \cite{Eaton2011SelectiveTB, Zuo2021AttentionBasedMD, Wang2022ClassrebalancedWD}. 

\section{Conclusion}
In this paper, we have presented Derivative Free Weight-space Ensembling (DFWE), a new few-sample task transfer approach. 
We showed how this approach can be effective in the open-domain dialogue setting.
Where proposing a conjunction of a finetuning and weight interpolation approach capable of transferring knowledge between many tasks in a way that the baseline fine-tuning approach cannot.
There are many ways to improve our approach, and we hope this serves to drive interest toward more efficient and effective transfer methods.

\bibliography{custom}


\end{document}